\documentclass[acmlarge, screen, nonacm]{acmart}

\setcopyright{none}
\renewcommand\footnotetextcopyrightpermission[1]{} % removes footnote with conference information in first column
\usepackage{url}
\usepackage{booktabs}
\usepackage{amsfonts}
\usepackage{nicefrac}
\usepackage{microtype}
\usepackage{amsmath}
\usepackage[justification=centering]{caption} 
\usepackage{multirow}
\usepackage{subcaption}
\usepackage{listings}
\usepackage{tcolorbox}
\usepackage{xspace}
\usepackage{makecell}
\usepackage[T1]{fontenc}
\usepackage[utf8]{inputenc}
\usepackage{hyperref}
\usepackage{enumitem}
\usepackage{booktabs}
\usepackage{multirow}
\usepackage{longtable}
\usepackage{pdflscape}
\usepackage{array}
\usepackage[normalem]{ulem}
\usepackage{setspace}
\useunder{\uline}{\ul}{}
\usepackage{tikz}
\usepackage[edges]{forest}
\usepackage{colortbl}
\definecolor{hidden-draw}{RGB}{20,68,106}
\definecolor{paired-dark-red}{RGB}{131, 60, 56}
\definecolor{bg22}{HTML}{B3E5FC}
\definecolor{pg53}{HTML}{B9F6CA} % light mint green
\definecolor{bg39}{HTML}{FFE0B2} % apricot
\definecolor{uli1}{HTML}{f4bf99}
\definecolor{highcolor}{RGB}{230, 255, 230}
\definecolor{mediumcolor}{RGB}{255, 255, 230}
\definecolor{lowcolor}{RGB}{255, 230, 230}

\tikzset{
    root style/.style={
        draw,
        rounded corners,
        fill=blue!30, % Color for the root node
        align=center,
        font=\bfseries
    },
    child style/.style={
        draw,
        rounded corners,
        fill=green!30, % Color for child nodes
        align=center,
        font=\bfseries
    },
    grandchild style/.style={
        draw,
        rounded corners,
        fill=red!30, % Color for grandchild nodes
        align=center,
        font=\bfseries
    }
}

\tikzset{
  my-box/.style={
    rectangle,
    draw=hidden-draw,
    rounded corners,
    text opacity=1,
    minimum height=1.5em,
    minimum width=40em,
    inner sep=2pt,
    align=center,
    line width=0.8pt,
  },
  leaf/.style={
    my-box,
    minimum height=1.5em,
    text=black,
    align=center,
    font=\normalsize,
    inner xsep=2pt,
    inner ysep=4pt,
    line width=0.8pt,
  }
}

\begin{document}
% \onehalfspacing

\title{Analysis of Indic Language Capabilities in LLMs}
% \author{Tattle Civic Tech}
% \email{admin@tattle.co.in}
\author{Aatman Vaidya}
\email{aatman@tattle.co.in}
\affiliation{%
  \institution{Tattle Civic Tech}
  \country{India}
}

\author{Tarunima Prabhakar}
\email{tarunima@tattle.co.in}
\affiliation{%
  \institution{Tattle Civic Tech}
  \country{India}
}

\author{Denny George}
\email{denny@tattle.co.in}
\affiliation{%
  \institution{Tattle Civic Tech}
  \country{India}
}

\author{Swair Shah}
\affiliation{%
    \country{India}
  % \institution{Tattle Civic Tech}
}

\thanks{*This work was funded by MLCommons to inform their work on AI Safety Benchmark, released as AI Luminate - \url{https://mlcommons.org/ailuminate/}}
\maketitle

\section{Introduction}
The rapid development in Large Language Models (LLMs) has resulted in state of the art performance on various natural language processing tasks such as text summarization, machine translation and sentiment analysis~\cite{zhao2024surveylargelanguagemodels, hendy2023good, McCandless_2024}. Large language models have become the dominant go-to approach towards building AI systems, of which many are deployed through conversational interfaces. Model developers claim that these models are multilingual. For example, OpenAI's GPT-4, GPT-4o, Anthropic's Claude and Meta's Llama family of models are popular models that were released as multilingual large language models. The technical reports of these models demonstrate their multilingual capabilities against benchmark datasets~\cite{metaIntroducingLlama, Mistral_Blog}. %For example, OpenAI claims that GTP-4o shows significant improvement in text in non-English languages and the language tokenization supports over 20 languages~\cite{gpt4o-report}. 
Some studies indicate that multilingual language models can understand connections between languages, allowing them to apply word associations and underlying grammatical rules learned from languages with more text data available to those with the lesser data~\cite{nicholas2023lost}. Yet, anecdotal experiences and other evaluations point to the limitations of multilingual performance of these models~\cite{Choudhary_2023a}. As we describe in Section 3.2, LLMs are trained on a large corpus of data crawled from the internet (e.g. C4 corpus, Books3 corpus etc.) where there is very sparse data for most spoken languages. This raises questions about their capabilities in low-resource languages~\cite{chatgpt-community-forum}. For instance, Microsoft Bing Chat, which relies on GPT-4, could not surface words for footwear in Spanish that are locally used in Latin America. Bing Chat also gave different and more vague answers, when the same question was translated from English to Spanish \cite{wired_chatgpt_non_english, Ali_2023}. In another analysis, a model, when pushed to do more complicated tasks like logical reasoning in Bengali it did not understand the instructions correctly. It also produced different results for Bengali and English~\cite{Deck_2023}. \\

This report provides an analysis of the capabilities of text-based LLMs to understand and generate Indic languages,\footnote{Languages spoken in the Indian Subcontinent} with the goal of providing suggestions on which Indian languages are best positioned for inclusion in future benchmarks datasets. 
% produced by MLCommons.
There are over a hundred languages spoken in the Indian subcontinent. More than forty of these have a million speakers but, only twenty-two languages are officially recognized in the Indian Constitution \cite{Rajbhasaha_Vibhag, lang_census_india}. The number of real-world speakers of a language however, does not directly correlate with the popularity of languages on the Internet. For example, Hindi- the language with the third highest number of speakers in the world- ranks at sixty-two in terms of number of Wikipedia articles in the language \cite{Dittus_Graham_2019}. This gap between language popularity online and in the real worlds bears on the performance and safety of LLMs that are predominantly trained on open access data.

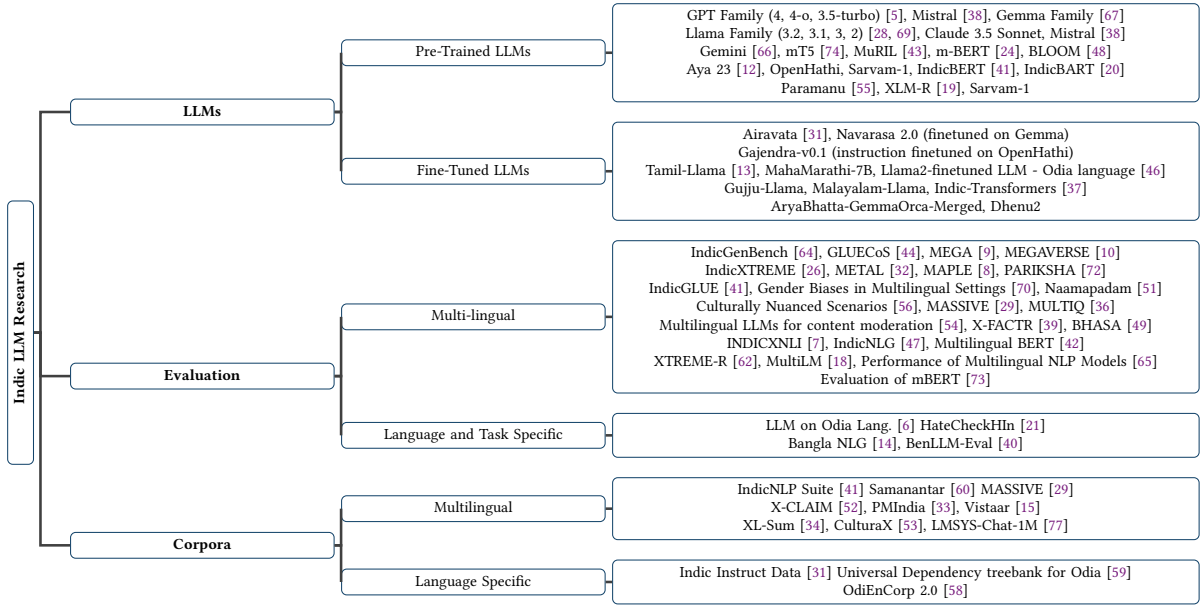
\begin{figure*}[ht!]
    \centering
    \resizebox{\textwidth}{!}{%
      \begin{forest}
        forked edges,
        for tree={
          grow=east,
          reversed=true,
          anchor=base west,
          parent anchor=east,
          child anchor=west,
          base=center,
          font=\Large,
          rectangle,
          draw=hidden-draw,
          rounded corners,
          align=center,
          text centered,
          minimum width=5em,
          text width=15em,
          edge+={darkgray, line width=2pt},
          s sep=15pt,
          l sep=5pt,
          inner xsep=2pt,
          inner ysep=3pt,
          line width=0.8pt,
          ver/.style={rotate=90, child anchor=north, parent anchor=south, anchor=center},
        },
        where level=0{
          rotate=90,
          text width=20em,
          anchor=south,
          parent anchor=south,
          child anchor=west,
          }{},
        where level=1{text width=20em,font=\Large}{},
        where level=2{text width=20em,font=\Large}{},
        where level=3{text width=45em,font=\Large}{},
        where level=4{text width=40em,font=\Large,}{},
        where level=5{text width=40em,font=\Large,}{},
        [
          \text{\textbf{Indic LLM Research}}
          [
           \text{\textbf{LLMs}}
           [
            \text{Pre-Trained LLMs}
            [
            {\text{GPT Family (4, 4-o, 3.5-turbo)} \cite{achiam2023gpt}, \text{Mistral} \cite{jiang2023mistral}, \text{Gemma Family} \cite{team2024gemma} \\ \text{Llama Family (3.2, 3.1, 3, 2)} \cite{dubey2024llama, touvron2023llama}, \text{Claude 3.5 Sonnet}, \text{Mistral} \cite{jiang2023mistral} \\ \text{Gemini} \cite{team2023gemini}, \text{mT5} \cite{xue2020mt5}, \text{MuRIL} \cite{khanuja2021muril}, \text{m-BERT} \cite{devlin2019bertpretrainingdeepbidirectional}, \text{BLOOM} \cite{le2023bloom} \\ \text{Aya 23} \cite{aryabumi2024aya23openweight}, \text{OpenHathi}, \text{Sarvam-1}, \text{IndicBERT} \cite{kakwani2020indicnlpsuite}, \text{IndicBART} \cite{dabre2021indicbart} \\ \text{Paramanu} \cite{niyogi2024paramanu}, \text{XLM-R} \cite{conneau2019unsupervised}, \text{Sarvam-1}}
            % \text{Llama Family (3.2, 3.1, 3, 2} \cite{dubey2024llama, touvron2023llama}
            ]
          ]
          [
          \text{Fine-Tuned LLMs}
          [
          {\text{Airavata} \cite{gala2024airavata}, \text{Navarasa 2.0 (finetuned on Gemma)} \\ \text{Gajendra-v0.1 (instruction finetuned on OpenHathi)} \\ \text{Tamil-Llama} \cite{balachandran2023tamil}, \text{MahaMarathi-7B}, \text{Llama2-finetuned LLM - Odia language} \cite{kohli2023building} \\ \text{Gujju-Llama}, \text{Malayalam-Llama}, \text{Indic-Transformers} \cite{jain2020indictransformersanalysistransformerlanguage} \\ \text{AryaBhatta-GemmaOrca-Merged, Dhenu2}}
          ]
          ]
          ]
          [
          \text{\textbf{Evaluation}}
          [
          \text{Multi-lingual}
          [
          {\text{IndicGenBench} \cite{singh2024indicgenbench}, \text{GLUECoS} \cite{khanuja2020gluecos}, \text{MEGA} \cite{ahuja2023mega}, \text{MEGAVERSE} \cite{ahuja2023megaverse} \\ \text{IndicXTREME} \cite{doddapaneni2022towards}, \text{METAL} \cite{hada2024metal}, \text{MAPLE} \cite{aggarwal2024maple}, \text{PARIKSHA} \cite{watts2024pariksha} \\ \text{IndicGLUE} \cite{kakwani2020indicnlpsuite}, \text{Gender Biases in Multilingual Settings} \cite{vashishtha2023evaluating}, \text{Naamapadam} \cite{mhaske2022naamapadam}  \\ \text{Culturally Nuanced Scenarios} \cite{ochieng2024beyond}, \text{MASSIVE} \cite{fitzgerald2022massive}, \text{MULTIQ} \cite{holtermann2024evaluating} \\ \text{Multilingual LLMs for content moderation} \cite{nicholas2023lost}, \text{X-FACTR} \cite{jiang2020xfactr}, \text{BHASA} \cite{leong2023bhasa} \\ \text{INDICXNLI} \cite{aggarwal2022indicxnli}, \text{IndicNLG} \cite{kumar2022indicnlg}, \text{Multilingual BERT} \cite{kassner2021multilingual} \\ \text{XTREME-R} \cite{ruder-etal-2021-xtreme}, \text{MultiLM} \cite{choudhury2021how}, \text{Performance of Multilingual NLP Models} \cite{srinivasan2021predicting} \\ \text{Evaluation of mBERT} \cite{wu-dredze-2020-languages} }
          ]
          ]
          [
          \text{Language and Task Specific}
          [
          {\text{LLM on Odia Lang.} \cite{10291329} \text{HateCheckHIn} \cite{das2022hatecheckhin} \\ \text{Bangla NLG} \cite{bhattacharjee-etal-2023-banglanlg}, \text{BenLLM-Eval} \cite{kabir2024benllmeval}}
          ]
          ]
          ]
          [
            \text{\textbf{Corpora}}
            [
            \text{Multilingual}
            [
            {\text{IndicNLP Suite} \cite{kakwani2020indicnlpsuite} \text{Samanantar} \cite{ramesh2022samanantar} \text{MASSIVE} \cite{fitzgerald2022massive} \\ \text{X-CLAIM} \cite{mittal2023lost}, \text{PMIndia} \cite{haddow2020pmindia}, \text{Vistaar} \cite{bhogale2023vistaar} \\ \text{XL-Sum} \cite{hasan2021xlsum}, \text{CulturaX} \cite{nguyen2023culturax}, \text{LMSYS-Chat-1M} \cite{zheng2023lmsys}}
            ]
            ]
            [
            \text{Language Specific}
            [
            {\text{Indic Instruct Data} \cite{gala2024airavata} \text{Universal Dependency treebank for Odia} \cite{parida2022universal} \\ \text{OdiEnCorp 2.0} \cite{parida-etal-2020-odiencorp}}
            ]
            ]
           ]
        ]
      \end{forest}
    }
    \caption{\textbf{Landscape of Indic AI Research.} (The structure and format has been sourced from KJ et al. \cite{kj2024decodingdiversityreviewindic})}
    \Description{a flowchart of indic ai research}
    \label{fig:lit_surv}
  \end{figure*}

Similar to \citet{kj2024decodingdiversityreviewindic}, we divide the analysis into three parts, to understand key elements of Indic AI research and development:
\begin{itemize}
    \item Analysis of existing LLMs and their Indic language capabilities. %% This includes models trained specifically for Indian languages, as well as an analysis of datasets used to commonly train multilingual large language models. 
    \item Summarization of existing evaluation studies and datasets that assess the capabilities of multilingual large language models (MLLMs) on specific language tasks.  
    
\end{itemize}
Figure~\ref{fig:lit_surv} presents a structured overview of the models, datasets, and evaluation methods considered in our analysis.  We conclude by contrasting LLM capabilities in Indic languages with real-world Indic language usage to provide suggestions on how to prioritize Indian languages for inclusion in future benchmarks. 

\section{Methodology}

%To survey the landscape of Indic AI, we develop a structured taxonomy based on key areas identified in previous reviews~\cite{qin2024multilingual, xu2024survey, kj2024decodingdiversityreviewindic, chang2024survey, peng2024survey}. Our analysis is organized into three main categories: \textbf{LLMs}, \textbf{Evaluation}, and \textbf{Corpora}. LLMs category focuses on the model themselves and has the sub-categories for pre-trained and fine-tuned LLMs. Evaluation category addresses the assessment methods and model performance across languages and tasks. Finally, Corpora looks at Indian language specific datasets.  The format has been sourced from . \\
This report has primarily been compiled through desk research. Since information about models is distributed across multiple sources, we scanned technical reports, announcement blogs, Hugging Face model cards, model research papers, survey papers and company press releases. For models that support Indian languages, we conduct a more detailed analysis using the Ecosystem Graphs framework proposed by ~\citet{bommasani2023ecosystem} to understand the following key attributes about the models:
\begin{itemize}
    \item \textbf{Host:} Organization that created the model
    \item \textbf{Model:} Name of the model
    \item \textbf{Date Created:} When the model was released
    \item \textbf{Size:} Number of parameters of the model
    \item \textbf{Training Dataset:} Dataset(s) (source) used to pre-train/fine-tune the model
    \item \textbf{Modality:} Input and Output modalities represented in the model, i.e. text, image, video, audio etc
    \item \textbf{Access:} Describes model accessibility. Types are Open (model weights and tokenizer available to users), Limited (paid API access), and Closed (no access).
    \item \textbf{License:} License of the model
    \item \textbf{LLM Type:} Pre-trained or Fine-tuned
    \item \textbf{Architecture:} Encoder only, Encoder-Decoder or Decoder only
\end{itemize}

While there are over 400 large language models \cite{McCandless_2024}, we could identify only twenty-eight models that are usable in Indian languages. Support for Indian languages was assessed from descriptions of data in technical report, description of model capabilities on official websites, or evaluation on multilingual benchmarks by the host organization. Models that claimed to be trained on multilingual corpus that contain Indian language data,\footnote{not necessarily optimized for state of the art of multilingual performance} or were evaluated on multilingual benchmarks were also included in our analysis. Where there are notable differences in models within a family of models released by an organization, such as LLama or Gemma, we include these as separate models in our analysis. We provide information on these attributes for 28 models in Table~\ref{tab:llm-list} in the Appendix. \\

%is selection was further guided by existing research in the Indian LLM space, where these models have been used in experiments or evaluated for various tasks.

The report also provides an overview of existing evaluation datasets and methodologies for assessing LLMs capabilities in Indian languages. We identified datasets by searching for papers on Google Scholar and ACL Anthology with keywords associated with Indian language evaluation, and from scanning references in notable papers on multilingual evaluation. We categorize these evaluation studies along three broad tasks, and their sub-tasks (See Figure~\ref{fig:eval-tasks}). This categorization was developed based on the tasks listed in existing literature on evaluation of LLMs for Indian languages~\cite{ahuja2023mega, ahuja2023megaverse}. We attempted to be comprehensive in the coverage of evaluations as they pertain to natural language generation and understanding, but not safety. 
%We detail the origins, architectures, training approaches and accessibility of the models in this section. Table \ref{tab:llm-list} summarizes attributes of these models such as their release dates, modalities, training datasets, access levels, licensing terms etc. The table presents an analysis of 30 prominent models that support Indian languages. The attributes for the information in the table have been sourced from Ecosystem Graphs project\footnote{\url{https://crfm.stanford.edu/ecosystem-graphs/index.html?mode=table}} by Stanford's CRFM Lab. 
%These attributes can help us understand the diversity, limitations, transparency, accessibility, usage and dependencies of LLMs~\cite{bommasani2023ecosystem}. Information for the models in Table~\ref{tab:llm-list} was gathered from various sources such as .

\section{Indic Language Capabilites in LLMs - An Overview}
% here is what the sections should be for LLM's
% 1. Propreity closed LLM'sd
% 2. Pre-trained LLM's
% 3. Fine-tuned LLM's
% Then talk about datsets that are used (corpoaa)
% after that talk about evaluation

% {\color{red} how to open this section? talk about figure 1 in opening statements.} 
Large language models are able to support Indic languages either because they are specifically designed to work on Indian languages through pre-training and fine-tuning, or because they are pre-trained on a multilingual corpus thereby supporting some Indian languages as a part of broader multilingual capabilities. For clarity, we call the former Indic language specific large language models (ILLMs), and the second multilingual large language models (MLLMs). From the over four hundred LLMs that have been released in the last three years, we identified only twenty-eight models that have some Indic language capabilities. These are listed in Table~\ref{tab:llm-list}. The twenty-eight models were developed by fourteen unique organizations. Half the models that we analyzed were developed by the US based corporations Meta, Google and OpenAI. Five models were developed by other AI corporations such as Mistral, Anthropic, and Cohere. Five models were developed by three India based corporations. Barring BLOOM which was developed by the Big Science consortium, the remaining were developed by research labs in India. Below we provide an overview of the capabilities of models in different Indic languages:

\subsection{Indic Language Specific Language Models}

Ten models of the twenty-eight we analyzed were specifically pre-trained or fine-tuned to work on an Indian language. Hindi, the most widely spoken Indic language, was covered by all the ten models. In addition to Hindi, most models also claim capabilities on nine other Indian languages. 

\begin{itemize}
   
    \item MuRIL by \cite{khanuja2021muril} is pre-trained on 17 Indian languages and their transliterated counterparts. This is developed by Google Research India. The model is pre-trained from scratch using data from sources like Wikipedia, Common Crawl, PMINDIA~\cite{haddow2020pmindia} and Dakshina~\cite{roark-etal-2020-processing}. 
    
    \item The Indian startup Sarvam has developed a number of models focused on Indic languages. Below we list the language coverage in their text based models:    
    \begin{itemize}
        \item OpenHathi model, is pre-trained from scratch in Hindi, English, and Indian English. However, as with many models we analyzed, information about the training data is not publicly available. 
        \item Sarvam 1 and its precursor Sarvam-2b-v0.5 is pretrained on language tokens for ten languages- Bengali, Gujarati, Hindi, Kannada, Malayalam, Marathi, Oriya, Punjabi, Tamil, and Telugu~\cite{Sarvam_2024}. Due to limited availability of good quality of Indic language data~\cite{kakwani2020indicnlpsuite}, the team developed a specialized training corpus of 2 trillion tokens using synthetic data-generation techniques. 
    \end{itemize}

    \item AryaBhatta-GemmaGenZ-Vikas-Merged, developed by GenVR Research, is finetuned on Gemma and supports nine Indian languages: Hindi, Tamil, Punjabi, Bengali, Gujarati, Oriya, Telugu, Kannada, Malayalam

    \item Open Aditi v4 and Gajendra-v0.1 are two other models that support Hindi and Hindi-English. Gajendra was finetuned on OpenHathi.

    \item Airavata by \citet{gala2024airavata} is finetuned on Indic Instruct dataset. This dataset is a collection of instruction datasets (Anudesh, wikiHow, Flan v2, Dolly, Anthropic-HHH, OpenAssistant v1, and LymSys-Chat).

\end{itemize}

\subsection{Multilingual Large Language Models}
Several models, while not specific about supporting Indian languages, do claim to be multilingual. These models have been evaluated for Indic language performance or have been used in building Indian languages specific applications. While information on training data is not available for many of these models,\footnote{Information on training data used was only available for seventeen of the twenty-eight models} analysing the Indic language composition of commonly used open access repositories in training provides some indication of the relative capabilities of MLLMs to support different Indic languages. Below we list the composition of Indian languages in some widely used training datesets:

\begin{itemize}
    \item Wikipedia: is an online encyclopedic repository in 332 languages. As an open access repository, data from Wikipedia is commonly used for training language models. In Table~\ref{tab:wiki-dist}, we present the percentage distribution of articles available in Indian languages on Wikipedia.
    
    \item Common Crawl: This is a collection of web archives consisting of terabytes of data. Common Crawl is a nonprofit organization that crawls the web and freely provides its archives and datasets to the public, it also includes data from Wikipedia. 
    While Common Crawl provides a multilingual dataset, the majority of its content is in English. Indian languages such as Hindi, Tamil, Bengali, Telugu, Marathi, and Gujarati comprise only about 0.5\% of the total data~\cite{common-crawl-lang}. In Table~\ref{tab:cc-dist} we list the percentage distribution of Indian languages in the Common Crawl dataset.

    \item Multilingual Colossal Clean Crawled Corpus (C4):
    This is a processed version of the Common Crawl created by Google, specifically for training MLLMs. It includes data for 101 languages of which Indian languages make up 7.72\% of the data~\cite{xue2021mt5massivelymultilingualpretrained}. 

\end{itemize}

\begin{table}[]
\begin{tabular}{lllll}
\hline
\textbf{Langauge} & \textbf{\begin{tabular}[c]{@{}l@{}}Percentage of \\ Articles (\%)\end{tabular}} &  & \textbf{Langauge}               & \textbf{\begin{tabular}[c]{@{}l@{}}Percentage of \\ Articles (\%)\end{tabular}} \\ \hline
Urdu              & 0.336                                                                             &  & Assamese                        & 0.022                                                                             \\
Tamil             & 0.265                                                                             &  & Maithili                        & 0.022                                                                             \\
Hindi             & 0.256                                                                             &  & Sanskrit                        & 0.019                                                                             \\
Bengali           & 0.25                                                                              &  & Santali                         & 0.018                                                                             \\
Telugu            & 0.159                                                                             &  & Bihari (Bhojpuri)               & 0.014                                                                             \\
Marathi           & 0.154                                                                             &  & Central Tibetan (Lhasa Tibetan) & 0.011                                                                             \\
Malayalam         & 0.135                                                                             &  & Kashmiri                        & 0.009                                                                             \\
Western Punjabi   & 0.115                                                                             &  & Konkani (Goan Konkani)          & 0.006                                                                             \\
Punjabi           & 0.086                                                                             &  & Pali                            & 0.004                                                                             \\
Kannada           & 0.052                                                                             &  & Awadhi                          & 0.004                                                                             \\
Gujarati          & 0.048                                                                             &  & Tulu                            & 0.004                                                                             \\
Odia              & 0.029                                                                             &  & Dzongkha                        & 0.00051                                                                           \\
Sindhi            & 0.029                                                                             &  &                                 &                                                                                   \\ \hline
                  &                                                                                   &  & \textbf{Total}                  & \textbf{2.047}                                                                    \\ \hline
\end{tabular}
\caption{Distribution of number of Articles in Indian Languages on Wikipedia}
\label{tab:wiki-dist}
\end{table}

% \begin{tabular}{cc}
% \hline
% \textbf{Langauge} & \textbf{Distribution (\%)} \\ \hline
% Hindi             & 0.1967                     \\
% Bengali           & 0.1075                     \\
% Tamil             & 0.0439                     \\
% Urdu              & 0.0298                     \\
% Malayalam         & 0.0244                     \\
% Marathi           & 0.0241                     \\
% Telugu            & 0.0181                     \\
% Kannada           & 0.0145                     \\
% Gujarati          & 0.0113                     \\
% Assamese          & 0.0023                     \\ \hline
% \textbf{Total}    & \textbf{0.4726}                     \\ \hline
% \end{tabular}
\begin{table}[]
\begin{tabular}{cclcc}
\hline
\textbf{Langauge} & \textbf{Percentage (\%)} &           & \textbf{Langauge} & \textbf{Distribution (\%)} \\ \hline
Hindi             & 0.1967                     &           & Marathi           & 0.0241                     \\
Bengali           & 0.1075                     &           & Telugu            & 0.0181                     \\
Tamil             & 0.0439                     &           & Kannada           & 0.0145                     \\
Urdu              & 0.0298                     &           & Gujarati          & 0.0113                     \\
Malayalam         & 0.0244                     &           & Assamese          & 0.0023                     \\ \hline
                  &                            & \textbf{} & \textbf{Total}    & \textbf{0.4726}            \\ \hline
\end{tabular}
\caption{Distribution of Indian Languages in the Common Crawl Dataset}
\label{tab:cc-dist}
\end{table}

Some organizations supplement these with additional open access repositories, in part, to build language specific capabilities. Here we note that Indic languages are usually not prioritized in these efforts. If an Indic language is selected at all, it is Hindi. For example, Llama 3.1 is trained on a custom dataset of 15T multilingual tokens from publicly available sources to support seven languages in addition to English, with Hindi being the only Indic language. Similarly, the Command R model by Cohere was updated with additional pre-training data for Hindi, among thirteen other global languages. Aya 23, which is an instruction fine-tuned on the Command model with support for 23 languages, also only covers Hindi among Indic languages.

\subsection{License and Access}
Our ability to analyze Indic language capabilities, as well as the ability of model and application developers to build in Indic languages is shaped by the license under which the models and dataset are released. 
All the models we analyzed were either open-access and limited-access models. \footnote{This refers strictly to ability to access, not the ability to inspect, modify or share the model} See Table ~\ref{tab:llm-list}.
For a majority of the models, model weights and tokenizers can be accessed through model cards on Hugging Face and GitHub. The source code and weights for Mistral, T5, mT5, Muril, Ganga 1B are available under permissive open-source licenses like Apache 2.0. The Cohere models are released under a Creative Commons license. The weight for Meta's Llama models are released under the Llama License and the code is released under GPLv3. Models like OpenHathi, Airavata, and Gajendra-v0.1 are fine-tuned on Llama, thereby inheriting its license. Similarly Navarasa 2.0 inherits its license from Gemma, the model it is finetuned on. Sarvam 1 model has been released under a custom Sarvam AI Research license. 
In Section 3.2 we describe the datasets used to train the models. We however note that barring a few exceptions such as MuRIL and Airavata, most models do not release their data under and an open data license.

\section{Evaluation Studies}

The overall performance of Large language models is evaluated by assessing their reliability, robustness, and effectiveness on a range of tasks. A comprehensive evaluation not only helps measure a model’s capabilities but also highlights areas that require further improvement. Popular LLMs like GPT and Llama report scores on well-rounded large benchmark datasets such as HELM, BIG-Bench, MMLU, MATH, and HellaSwag etc, which provide insights into their general performance across multiple domains~\cite{gpt4o-report}. These benchmarks however don't evaluate language specific performance, and also don't account for context specific usage of LLMs. This has led some groups to develop Indic language specific evaluation datasets %However, when LLMs are required to perform specific tasks, the existence of evaluation methods tailored to those tasks becomes helpful. 
In this section, we summarize existing evaluation datasets and methods for Indic language performance of LLMs. Please note that the LLMs covered in the evaluation papers may differ from the 28 models analyzed in the previous section. \\

%With the growing evolution in LLMs, refined methods to evaluate the capabilities of LLMs are needed to determine the tasks and responsibilities they should undertake. 

\begin{figure*}[]
    \centering
    \includegraphics[width=\linewidth]{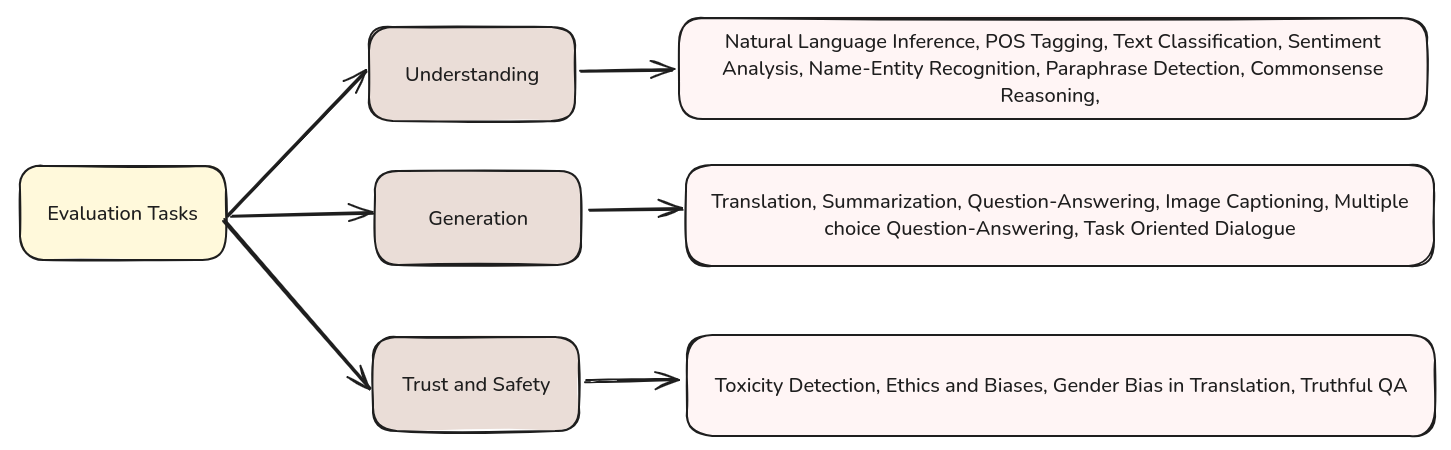}
    \caption{Evaluation Tasks}
    \label{fig:eval-tasks}
\end{figure*}

To evaluate LLMs effectively, it is essential to assess their core abilities, specifically their capacity to understand and generate natural language text. This ability is assessed through a number of smaller tasks. In Figure~\ref{fig:eval-tasks} we present a small taxonomy of evaluation tasks and their sub-tasks. As illustrated, the evaluation tasks can be broadly categorized into three main domains: \textbf{Natural Language Understanding}, \textbf{Natural Language Generation}, and \textbf{Safety}, each with multiple sub-tasks:

\begin{itemize}
    \item \textbf{Natural Language Understanding} refers to LLMs' ability to comprehend and interpret human language. It involves processing the input text to extract meaning, context, and underlying relationships between words. There are several tasks that fall under this, such as Natural Language Inference, Text Classification, Sentiment Analysis, Name-Entity Recognition (NER)\footnote{NER is a process of identifying and classifying named entities (people, objects, locations) in text} and POS Tagging \footnote{POS tagging is a way to tag different parts of speech such as noun, verb, adjective, pronoun, prepositions in an given text.}.
    \item \textbf{Natural Language Generation} is the ability of LLMs' to generate text based on a given input. Some key tasks that fall under it are Translation, which is a models ability to convert text from one language to the other, Summarization, Question-Answering, Image Captioning etc.
    \item \textbf{Safety} focuses on the crucial evaluation of the ethical and safety aspects of LLMs. 
\end{itemize}  

Since the goal of the report is to provide suggestions on which Indic language should be included in future safety benchmark datasets, we preclude an analysis of trust and safety datasets relevant to Indic languages. The following is a summarization of model performance on natural language understanding and generation.

% \subsection{Tasks Included in Evaluations}
% Table~\ref{tab:eval-dataset-lang-covered} summarizes the languages covered by existing evaluation benchmarks.

\subsection{Summary of Model Evaluations}

We found fourteen papers that evaluated an LLM against one or more Indic languages. A detailed description of the papers can be found in \textbf{Section~\ref{sec:appendix-eval-papers}} of the Appendix. We note that a majority of these datasets are a direct translation of existing English datasets which is a serious limitation. For example, while models might score well on some of these datasets, they may fare poorly on tasks where socio-cultural understanding is required~\cite{verma2024milumultitaskindiclanguage}. The following analysis should be read alongside a recognition of this limitation in existing evaluation datasets. 

\begin{itemize}

    \item \textbf{Language Distribution in Evaluation Datasets:}~In Table~\ref{tab:eval-dataset-lang-covered} we list the languages covered in existing evaluation studies across tasks. Except for the work by \citet{singh2024indicgenbench}, we observed that all other datasets primarily support a maximum of the 12 most widely spoken Indian languages.
    \begin{itemize}
        \item These are Hindi, Bengali, Gujarati, Kannada, Malayalam, Marathi, Tamil, Telugu, Urdu, Assamese, Odia and Punjabi.
    \end{itemize}
          
    \item \textbf{Performance of LLMs:}~To understand the relative capabilities of the models in Indic languages, we categorize the twelve languages in three broad buckets of \textsc{high}, \textsc{medium} and \textsc{low}
    (See Table~\ref{tab:eval-perform-across-lang}). \textbf{This ranking should not be treated as a universal ranking- there is a significant performance gap in LLMs between English and Indian languages on natural language understanding and generation tasks. The rankings in Table~\ref{tab:eval-perform-across-lang} are relative to each other.} 
    
    While it is difficult to summarize the performance of languages across all evaluation datasets, models and sub-tasks, we provide a high-level approximation of the performance of the models on natural language understanding and generation tasks. A language was labeled as \textsc{high} performing on a task if it consistently ranked at the top or had scores similar to the highest ranked language, across a majority of models. A language was labeled as \textsc{medium} performing, if it had average performance on most sub-tasks, on a majority models. A language was labeled as \textsc{low} performing, if it has poor scores across most sub-tasks on a majority of models. We identify that LLMs shows relatively \textsc{high} performance on six Indic languages, \textsc{medium} on four and \textsc{low} on two of the twelve. We also note that:
    
    \begin{itemize}
        \item The performance gap between high and low resource languages is because of a lack of good quality and quantity of pre-training data. This has led some developers to create and use synthetic data.
        \item \citet{doddapaneni2023leaving} in reviewing performance of LLMs on twenty Indian languages notice a sharp decline in performance of extremely low resource languages such as Sindhi and Santhali. They also attribute this performance gap to the lack of a shared script among languages, and absence of linguistic cousin in the corpus to act as a bridge for effective transfer.
       
    \end{itemize}
    % talk about size of the dataest also, talk about tokenization fertility
   
\end{itemize}

\begin{table}[]
\resizebox{\textwidth}{!}{
\begin{tabular}{llll}
\hline
\multicolumn{1}{c}{\textbf{Tasks}}                                                         & \multicolumn{1}{c}{\textbf{Sub-Tasks}}                                & \multicolumn{1}{c}{\textbf{Languages Covered}}                                                                                                                                                                                                                                                                          & \multicolumn{1}{c}{\textbf{Papers Referenced}}                                                                                                                                                                                             \\ \hline
\multirow{3}{*}{\begin{tabular}[c]{@{}l@{}}Natural Language \\ Understanding\end{tabular}} & \begin{tabular}[c]{@{}l@{}}Natural Language \\ Inference\end{tabular} & \begin{tabular}[c]{@{}l@{}}Hindi, Bengali, Punjabi, Kannada, Gujarati, Malayalam, \\ Marathi, Telugu, Tamil, Oriya, \\ Assamese, code-mixed English-Hindi, Nepali\end{tabular}                                                                                                                                          & \begin{tabular}[c]{@{}l@{}}\citet{ahuja2023mega}, \\ \citet{doddapaneni2022towards}, \\ \citet{aggarwal2022indicxnli}\end{tabular}                                                 \\ \cline{2-4} 
                                                                                           & Text Classification                                                   & \begin{tabular}[c]{@{}l@{}}Hindi, Kannada, Malayalam, Marathi, Tamil, Telugu, \\ Urdu, Bengali, Gujarati,\\ Assamese, Odia, Punjabi\end{tabular}                                                                                                                                                                        & \begin{tabular}[c]{@{}l@{}}\citet{doddapaneni2022towards}, \\ \citet{kakwani2020indicnlpsuite}\end{tabular}                                                                                               \\ \cline{2-4} 
                                                                                           & \begin{tabular}[c]{@{}l@{}}Name-Entity \\ Recognition\end{tabular}    & \begin{tabular}[c]{@{}l@{}}Urdu, Tamil, Telugu, Hindi, Gujarati, Malayalam, \\ Marathi, Punjabi, Bengali, Kannada\end{tabular}                                                                                                                                                                                          & \begin{tabular}[c]{@{}l@{}}\citet{ahuja2023mega}, \\ \citet{doddapaneni2022towards},\\ \citet{kakwani2020indicnlpsuite}\end{tabular}                                                     \\ \hline
\multirow{4}{*}{\begin{tabular}[c]{@{}l@{}}Natural Language \\ Generation\end{tabular}}    & Translation                                                           & \begin{tabular}[c]{@{}l@{}}Bengali, Gujarati, Hindi, Kannada, Malayalam, Marathi, \\ Tamil, Telugu, Urdu, Assamese, Bhojpuri, Nepali, \\ Odia, Punjabi, Pashto, Sanskrit, Awadhi, Haryanvi, Tibetan, \\ Bodo, Garhwali, Konkani, Chhattisgarhi, Rajasthani, \\ Maithili, Manipuri, Malvi, Marwari, Santali\end{tabular} & \citet{singh2024indicgenbench}                                                                                                                                                                                            \\ \cline{2-4} 
                                                                                           & Summarization                                                         & \begin{tabular}[c]{@{}l@{}}Bengali, Gujarati, Hindi, Kannada, Malayalam, Marathi, \\ Tamil, Telugu, Urdu, Assamese, Bhojpuri, Nepali, \\ Odia, Punjabi, Pashto, Sanskrit, Awadhi, Haryanvi, Tibetan, \\ Bodo, Garhwali, Konkani, Chhattisgarhi, Rajasthani, \\ Maithili, Manipuri, Malvi, Marwari, Santali\end{tabular} & \begin{tabular}[c]{@{}l@{}}\citet{singh2024indicgenbench}, \\ \citet{ahuja2023mega},\\ \citet{hada2024metal}, \\ \citet{kumar2022indicnlg}\end{tabular}          \\ \cline{2-4} 
                                                                                           & Question-Answering                                                    & \begin{tabular}[c]{@{}l@{}}Bengali, Gujarati, Hindi, Kannada, Malayalam, \\ Marathi, Tamil, Telugu, Urdu, Assamese, Odia, Punjabi\end{tabular}                                                                                                                                                                          & \begin{tabular}[c]{@{}l@{}}\citet{singh2024indicgenbench}, \\ \citet{ahuja2023mega}, \\ \citet{doddapaneni2022towards},\\ \citet{kakwani2020indicnlpsuite}\end{tabular} \\ \cline{2-4} 
                                                                                           & Image Captioning                                                      & Bengali, Hindi, Telugu                                                                                                                                                                                                                                                                                                  & \citet{ahuja2023megaverse}                                                                                                                                                                                                \\ \hline
       
\end{tabular}
}
\caption{Tasks and Languages Covered in Existing Evaluation Datasets}
\label{tab:eval-dataset-lang-covered}
\end{table}

\begin{table}[]
\small
\resizebox{\textwidth}{!}{
\begin{tabular}{lcccc}
\hline
\multicolumn{1}{c}{\multirow{2}{*}{\textbf{Langauge}}} & \multicolumn{3}{c}{\textbf{Performance of LLM on Tasks}}                                                                                                                                                                                                                               & \multirow{2}{*}{\textbf{\begin{tabular}[c]{@{}c@{}}Number of\\ Speakers in India\end{tabular}}} \\ \cline{2-4}
\multicolumn{1}{c}{}                                   & Understanding & Generation & \begin{tabular}[c]{@{}c@{}}Evaluation Studies \\ Referenced\end{tabular}                                                                                                                                                       &                                                                                                 \\ \hline
Hindi                                                  & \textsc{high}                   & \textsc{high}                & \multirow{6}{*}{\begin{tabular}[c]{@{}c@{}}\citet{singh2024indicgenbench}, \citet{ahuja2023mega}, \citet{aggarwal2022indicxnli}, \\ \citet{doddapaneni2022towards}, \citet{kakwani2020indicnlpsuite}\end{tabular}} & 528,347,193                                                                                     \\ \cline{1-3} \cline{5-5}
Bengali                                                & \textsc{high}                   & \textsc{high}                &                                                                                                                                                                                                                                         & 97,237,669                                                                                      \\ \cline{1-3} \cline{5-5}
Marathi                                                & \textsc{high}                   & \textsc{high}                &                                                                                                                                                                                                                                         & 83,026,680                                                                                      \\ \cline{1-3} \cline{5-5}
Telugu                                                 & \textsc{high}                   & \textsc{high}                &                                                                                                                                                                                                                                         & 81,127,740                                                                                      \\ \cline{1-3} \cline{5-5}
Tamil                                                  & \textsc{high}                   & \textsc{high}               &                                                                                                                                                                                                                                         & 69,026,881                                                                                      \\ \cline{1-3} \cline{5-5}
Gujarati                                               & \textsc{medium}                 & \textsc{medium}              &                                                                                                                                                                                                                                         & 55,492,554                                                                                      \\ \hline
Urdu                                                   & \textsc{high}                   & \textsc{high}                & \citet{singh2024indicgenbench}, \citet{ahuja2023mega}, \citet{doddapaneni2022towards}                                                                                                                                  & 50,772,631                                                                                      \\ \hline
Kannada                                                & \textsc{medium}                 & \textsc{medium}              & \multirow{6}{*}{\begin{tabular}[c]{@{}c@{}}\citet{singh2024indicgenbench}, \citet{ahuja2023mega}, \citet{aggarwal2022indicxnli}, \\ \citet{doddapaneni2022towards}, \citet{kakwani2020indicnlpsuite}\end{tabular}} & 43,706,512                                                                                      \\ \cline{1-3} \cline{5-5}
Oriya                                                  & \textsc{low}                    & \textsc{low}                 &                                                                                                                                                                                                                                         & 37,521,324                                                                                      \\ \cline{1-3} \cline{5-5}
Malayalam                                              & \textsc{medium}                 & \textsc{medium}              &                                                                                                                                                                                                                                         & 34,838,819                                                                                      \\ \cline{1-3} \cline{5-5}
Punjabi                                                & \textsc{low}                    & \textsc{low}                 &                                                                                                                                                                                                                                         & 33,124,726                                                                                      \\ \cline{1-3} \cline{5-5}
Assamese                                               & \textsc{medium}                 & \textsc{medium}             &                                                                                                                                                                                                                                         & 15,311,351                                                                                      \\ \hline 
\end{tabular}
}
\caption{Overall Performance of LLMs on Evaluation Tasks for Indian Languages}
\label{tab:eval-perform-across-lang}
\end{table}

% see section these papers for trust and safety datasets - \href{https://arxiv.org/pdf/2406.00936}{paper1}, \href{https://dl.acm.org/doi/pdf/10.1145/3641289#page=3.24}{paper2}, also include indic gen by ai4bharat - \href{https://huggingface.co/datasets/ai4bharat/IN22-Gen}{link} and the latests eval paper

% \section{Corpora}

\section{Prioritizing Languages for Inclusion in Benchmarks}
The goal for this report was to provide an overview of Indic language capabilities of prominent LLMs and to suggest languages that should be included in future safety benchmarks. The prior sections provide an overview of the technical capabilities of the models. We note, however, that the performance of models in a language is not directly correlated with the number of people who speak the language. While models show good language generation and understanding in the five most widely spoken languages in India- Hindi, Bengali, Marathi, Telugu and Tamil- the performance for other languages is less predictable. As Table \ref{tab:eval-perform-across-lang} shows, Oriya while having more speakers than Malayalam and Assamese, performs worse. 

Based on Table \ref{tab:eval-perform-across-lang} Hindi, Bengali, Marathi, Telugu and Tamil emerge as natural candidates for inclusion in safety benchmarks.\footnote{Hindi is already included in v1 AI safety benchmark released by MLCommons} For these languages, the model capabilities dovetail the number of language speakers in the country. Beyond this, however, the selection criteria becomes more complicated. The performance of the model in a language may be low. Yet, the number of people using the model in the language may be higher and therefore repercussion from unsafe usage might be greater.\footnote{This gap between language speakers and model capabilities can be addressed by model developers but is out of scope for the MLCommons' AI Safety Working Group}. Model performance on specific languages might also rapidly change with the availability of newer datasets in the language. At this stage we find it difficult to unequivocally recommend languages beyond the five most widely spoken in India.

% The decision for selecting specific languages, driven by resource constraints, can be made on several criteria:
% \begin{itemize}
%     \item Indic Languages that are already supported by LLMs should be prioritized first.
%     \item Languages should be prioritized based on the number of real world speakers
%     \item Languages should be prioritized based on known correlated risks. For example a language, even if not the most widely spoken, it should be prioritized if the population speaking it is at risk of violence.\footnote{This is the principle that justifies prioritizing effective moderation in languages such as Burmese even though it is spoken by less than fifty million people} 
% \end{itemize}

\begin{acks}
This report provides an analysis of the capabilities of text-based LLMs to understand and generate Indic languages with the goal of providing suggestions on which Indian languages are best positioned for inclusion in future benchmarks datasets by MLCommons. This work is supported and funded by MLCommons. 
\end{acks}

\bibliographystyle{ACM-Reference-Format}
\bibliography{sample-base}

\appendix

\section{Indic LLM Table Data}

\begin{longtable}{>{\centering\arraybackslash}m{1.5cm}   % Host
                  >{\centering\arraybackslash}m{1.3cm}   % Model
                  >{\centering\arraybackslash}m{1.2cm}   % Created Date
                  >{\centering\arraybackslash}m{1cm}     % Size
                  >{\centering\arraybackslash}m{2.5cm}   % Dataset
                  >{\centering\arraybackslash}m{1.5cm}   % Modality
                  >{\centering\arraybackslash}m{1cm}     % Access
                  >{\centering\arraybackslash}m{1.3cm}   % License
                  >{\centering\arraybackslash}m{1.3cm}   % Type
                  >{\centering\arraybackslash}m{1cm}}    % Architecture

\caption{An Overview of Indic LLM's, including their release time, modality, access, license, training data and size}\label{tab:llm-list}\\
\hline
\textbf{Host} & \textbf{Model} & \textbf{Date Created} & \textbf{Size} & \textbf{Training Dataset} & \textbf{Modality (In; Out)} & \textbf{Access} & \textbf{License} & \textbf{LLM Type} & \textbf{Architecture} \\ \hline
\endfirsthead

\multicolumn{10}{c}%
{{\bfseries \tablename\ \thetable{} -- continued from previous page}} \\
\hline
\textbf{Host} & \textbf{Model} & \textbf{Date Created} & \textbf{Size} & \textbf{Training Dataset} & \textbf{Modality (In; Out)} & \textbf{Access} & \textbf{License} & \textbf{LLM Type} & \textbf{Architecture} \\ \hline
\endhead

% \hline \multicolumn{10}{|r|}{{Continued on next page}} \\ \hline
% \endfoot

\hline
\endlastfoot

\multirow{4}{*}{OpenAI}                 & GPT 4                             & 14 March 2023         & unknown       & -                                                                                                          & image, text; image, text                      & limited         & unknown                                     & pre-trained       & decoder only          \\ \cline{2-10} 
                                        & GPT 3.5-turbo                     & 1 March 2023          & unknown       & -                                                                                                          & text; text                                    & limited         & Custom                                      & pre-trained       & decoder only          \\ \cline{2-10} 
                                        & GPT 4-o                           & 13 May 2024           & unknown       & -                                                                                                          & audio, image, text, video; audio, image, text & limited         & unknown                                     & pre-trained       & decoder only          \\ \cline{2-10} 
                                        & GPT 3                             & 11 June 2020          & 175B          & Common Crawl, WebText dataset, Books corpora and English-language Wikipedia data                           & text; text                                    & limited         & unknown                                     & pre-trained       & decoder only          \\ \hline
Anthropic                               & Claude 3.5 Sonnet                 & 21 June 2021          & unknown       & -                                                                                                          & text; image, text                             & open            & unknown                                     & pre-trained       & decoder only          \\ \hline
Mistral                                 & Mistral                           & 27 Sep 2023           & 7B            & -                                                                                                          & text; text                                    & open            & Apache 2.0                                  & pre-trained       & decoder only          \\ \hline
\multirow{6}{*}{Google}                 & Gemma                             & 21 Feb 2024           & 7B            & -                                                                                                          & text; text                                    & open            & Custom                                      & pre-trained       & decoder only          \\ \cline{2-10} 
                                        & Gemma 2                           & 21 Feb 2024           & 27B           & -                                                                                                          & text; text                                    & open            & Custom                                      & pre-trained       & decoder only          \\ \cline{2-10} 
                                        & Gemini 1.5 Flash                  & 30 May 2024           & unknown       & -                                                                                                          & audio, image, text, video; text               & limited         & Googles Terms and Conditions                & pre-trained       & decoder only          \\ \cline{2-10} 
                                        & mT5                               & 22 Oct 2020           & 11B (large)   & Multilingual C4 (mC4), it has 101 languages and is generated from 86 Common Crawl dumps                    & text; text                                    & open            & Apache 2.0                                  & pre-trained       & encoder-decoder       \\ \cline{2-10} 
                                        & Muril                             & 19 Mar 2021           & -             & Wikipedia and Common Crawl for 17 Indian languages, includes translated and transliterated data            & text; text                                    & open            & Apache 2.0                                  & pre-trained       & encoder only          \\ \cline{2-10} 
                                        & mBERT                             & 5 Nov 2018            & -             & 104 languages with the highest Wikipedia data                                                              & text; text                                    & open            & Apache 2.0                                  & pre-trained       & encoder only          \\ \hline
\multirow{4}{*}{Meta}                   & Llama 3                           & 18 Apr 2024           & 70B           & Trained on over 15 trillion tokens of data from publicly available sources.                                & text; text                                    & open            & Llama 3 Community License Agreement         & pre-trained       & decoder only          \\ \cline{2-10} 
                                        & Llama                             & 24 Feb 2023           & 65B           & CommonCrawl, C4, Github, Wikipedia, BooksCorpus, arXiv, StackExchange                                      & text; text                                    & open            & LLaMa License (model weights), GPLv3 (code) & pre-trained       & decoder only          \\ \cline{2-10} 
                                        & Llama2                            & 18 Jul 2023           & 7B, 70B       & Trained on a new mix of publicly available data.                                                           & text; text                                    & open            & Llama 2 Community License Agreement         & pre-trained       & decoder only          \\ \cline{2-10} 
                                        & Llama 3.1                         & 23 Jul 2024           & 8B, 70B, 405B & Trained on over 15 trillion tokens of data from publicly available sources.                                & text; text                                    & open            & Llama 3.1 Community License Agreement       & pre-trained       & decoder only          \\ \hline
Big Science                             & BLOOM                             & 12 Jul 2022           & 176B          & ROOTS corpus                                                                                               & text; text                                    & open            & BigScience RAIL License                     & pre-trained       & decoder only          \\ \hline
\multirow{2}{*}{Cohere}                 & Command R                         & 11 Mar 2024           & 35B           & -                                                                                                          & text; text                                    & open            & CC BY NC 4.0                                & pre-trained       & decoder only          \\ \cline{2-10} 
                                        & Aya-23                            & 31 May 2024           & 35B           & Aya Dataset - human-curated prompt-completion pairs                                                        & text; text                                    & open            & CC BY NC 4.0                                & finetuned         & decoder only          \\ \hline
\multirow{3}{*}{Sarvam AI}              & OpenHathi                         & -                     & 7B            & -                                                                                                          & text; text                                    & open            & Llama 2 Community License Agreement         & pre-trained       & decoder only          \\ \cline{2-10} 
                                        & Sarvam-2b-v0.5                    & 13 Aug 2024           & 2B            & Trained on a data mixture of 4 trillion tokens: containing equal parts English (2T) and Indic (2T) tokens. & text; text                                    & open            & unkown                                      & pre-trained       & decoder only          \\ \cline{2-10} 
                                        & Sarvam 1                          & 24 Oct 2024           & 2B            & Sarvam-2T - includes 2T tokens for 10 Indic language                                                       & text; text                                    & open            & Sarvam AI Research License                  & pre-trained       & decoder only          \\ \hline
AI4Bharat                               & Airavata                          & 26 Jan 2024           &               & Indic Instruct Data                                                                                        & text; text                                    & open            & Llama 2 Community License Agreement         & finetuned         & decoder only          \\ \hline
Telugu LLM Labs                         & Navarasa 2.0                      & -                     & 2B, 7B        & LoRA finetuned on 15 Indian languages and English language instruction datasets:                           & text; text                                    & open            & Gemma Terms of Use                          & finetuned         & decoder only          \\ \hline
Lingo Research Group at IIT Gandhinagar & Ganga 1B                          & -                     & 1B            & trained on a large dataset of public domain web-crawled Hindi language data                                & text; text                                    & open            & Apache 2.0                                  & finetuned         & decoder only          \\ \hline
Bhabha AI                               & Gajendra-v0.1                     & -                     & 7B            & -                                                                                                          & text; text                                    & open            & Llama 2 Community License Agreement         & finetuned         & decoder only          \\ \hline
-                                       & Open Aditi v4                     & -                     & 7B            & Samvadd                                                                                                    & text; text                                    & open            & Apache 2.0                                  & finetuned         & decoder only          \\ \hline
GenVR Research                          & AryaBhatta-GemmaGenZ-Vikas-Merged & -                     & 7B            & GenZ Vikas dataset                                                                                         & text; text                                    & open            & MIT                                         & finetuned         & decoder only          \\ \hline
\end{longtable}

\section{Indic LLM Evaluation Papers}
\label{sec:appendix-eval-papers}

In this section, we review relevant papers that evaluate large language models for Indian languages, including the models assessed, the datasets created/ used, languages and tasks covered and the methodologies employed.
\begin{itemize}
    \item IndicGenBench by~\citet{singh2024indicgenbench} extends existing benchmarks to 29 Indic languages through human annotation. The dataset comprises of diverse generation tasks like cross-lingual summarization, machine translation, and cross-lingual question answering. The authors evaluate a wide range of proprietary and open-source LLMs like GPT-3.5, GPT-4, PaLM-2, mT5, Gemma, BLOOM and LLaMA. The work finds a significant performance gap in all languages compared to English. They also compares the token fertility (average number of sub-words that a word is broken down into by the tokenizer) across all Indic languages in the dataset. They find that Urdu has the lowest token fertility and Tibetan has the highest.
    \item MEGA by~\citet{ahuja2023mega} presents a comprehensive benchmarking of generative LLMs for 70 typologically diverse languages across 16 NLP datasets. The authors compare the performance of LLMs like GPT-3.5, GPT-4, BLOOMZ, m-BERT. They find a significant performance gap between English and non-English languages, especially low-resource languages with non-Latin scripts. Three of the 16 datasets focus exclusively on Indian languages, covering over 11 Indic languages, while other datasets include some Indic languages as well. The authors also explore different prompting strategies, finding that translating test data to English before inputting it to the model often improves performance for low-resource languages. They create a framework for for evaluating generative LLMs in the multilingual setting and provide directions for future research.
    \item MEGAVERSE by~\citet{ahuja2023megaverse} extends MEGA coverage to 22 datasets and 83 languages (including many low-resource African languages). The dataset suite includes tasks such as text classification, POS tagging, NER, translation, summarization, question answering etc. They benchmark performance of GPT-3.5-Turbo, GPT-4, PaLM2, GeminiPro, Mistral, Llama2, Gemma, and multimodal models like LLaVA, GPT-4-Vision, and GeminiPro-Vision. Results show that larger models, particularly GPT-4, Gemini-Pro, and PaLM2, outperform smaller models, especially on low-resource languages. The authors also conduct a data contamination study and point out that several models are likely contaminated with multilingual evaluation benchmarks, suggesting the need for approaches to detect and handle contamination when assessing the multilingual performance of LLMs.
    \item IndicXTREME by~\citet{doddapaneni2022towards} is a human supervised benchmark containing evaluation sets for nine diverse tasks with each task covering 7-18 Indic languages per task. These are five sentence classification tasks, two structure prediction, one question answering and one sentence retrieval task. The authors evaluate IndicBERT, mBERT, XLMR, and MuRIL on the dataset, showing that pre-training on Indic languages enhances performance, with additional gains from using in-language or related language data for training and development.
    \item METAL by~\citet{hada2024metal} contributes a datasets to evaluate LLM-based evaluators, which they refer to as meta-evaluation. The task covered in the dataset is summarization and contains 10 languages, out of Hindi and Bengali are the Indic languages. The authors evaluate the performance of GPT-3.5, GPT-4 and PaLM-2 on the dataset and results show that GPT-4 with detailed instructions performs closest to humans, while GPT-3.5-Turbo is not a suitable multilingual evaluator.
    \item PARIKSHA by~\citet{watts2024pariksha} studies human and LLM-based evaluation in a multilingual and multi-cultural setting. They evaluate 30 models across 10 Indic languages. The task was to evaluate the output of LLMs based on prompts, covering topics in finance, health, and culture. They find that models such as GPT-4o and Llama-3 70B consistently perform best for most Indic languages on pairwise settings and human evaluation. Their findings also indicate that LLM evaluators tend to align less with human evaluations on assessing culturally nuanced responses.
    \item IndicNLPSuite by~\citet{kakwani2020indicnlpsuite} is a collection of resources for Indian language NLP. In this this suite, they introduce IndicGLUE, a benchmark dataset spanning several natural language understanding tasks supporting 11 Indic languages. The tasks include News Category Classification, Headline Prediction, Wikipedia Section-title Prediction, Multiple-choice QA, NER and Cross-lingual Sentence Retrieval. They evaluate XLM-R, mBERT, IndicBERT (large and base) on the dataset and IndicBERT performed the best on most tasks. 
    \item IndicNLG by~\citet{kumar2022indicnlg} is a collection of natural language generation datasets for 11 Indic languages. The tasks include biography generation using Wikipedia infoboxes, news headline generation, sentence summarization, generation and, question generation. They evaluate two models, IndicBART and mT5, and find that language-specific models like IndicBART demonstrate better performance, highlighting the advantage of pre-training on language-specific data.
    \item IndicXNLI by~\citet{aggarwal2022indicxnli} is a natural language inference datasets supporting 11 Indic languages. The datasets is created by machine translating the original English XNLI dataset. The dataset’s quality is validated through human evaluation. The authors fine-tune XLM-R, IndicBERT, mBERT, and MuRIL on the dataset and investigate cross-lingual transfer techniques. Results show that MuRIL generally performs best, and training on both English and Indic data yields the highest scores.
    \item MILU (Multi-task Indic Language Understanding) by~\citet{verma2024milumultitaskindiclanguage} is benchmark dataset covering 11 Indic languages. The dataset includes tasks for natural language understanding and generation, evaluating both linguistic competence and cultural understanding. It spans 8 domains and 42 subjects and is designed with an India-first perspective by collecting questions from various national, state, and regional exams. These questions include culturally relevant subjects such as local history, arts, festivals, and laws, alongside traditional academic subjects like science and mathematics. The authors evaluate 45 different LLMs (mix of closed proprietary, open-source, and language-specific models), and the findings suggest that models struggle with MILU, with GPT-4o achieving the highest average accuracy at 72\%. Interestingly, open multilingual models outperform language specific models. Their domain-wise analysis reveals that models perform poorly in culturally relevant areas, such as Arts \& Humanities and Social Sciences, compared to more general fields like STEM.
    \item MAPLE by~\citet{aggarwal2024maple} focuses on multilingual evaluation of parameter efficient fine-tuning of LLMs. They analyze the effects of percentage of trainable parameters and quantization on 40 languages across various tasks like classification, question answering, summarization and institution following. The paper compares performance of Llama-2-7B and Mistral-7B on the datasets. They find that having more languages in the fine-tuning datasets does not necessarily mean significantly better multilingual performance.
    % \item GLUECoS by~\cite{khanuja2020gluecos} presents an evaluation benchmark for code-switched Indian English language. The tasks included are Language Identification from text, POS tagging, Named Entity Recognition, Sentiment Analysis and Question Answering.
    \item Indic LLM Leaderboard translates existing SOTA benchmarks like benchmarks like ARC, Hellaswag, MMLU, etc to 7 Indic languages. Maintains a leader board evaluating pre-trained and fine-tuned Indic LLMs.
\end{itemize}







\end{document}